\documentclass[letterpaper]{article}
\usepackage{proceed2e}
\usepackage[margin=1in]{geometry}

\usepackage{amsmath,amssymb}
\usepackage{hyperref}

\usepackage{graphicx}
\graphicspath{ {images/} }

\usepackage{acronym}
\usepackage{etoolbox}
\preto\section\acresetall

\acrodef{rl}[RL]{Reinforcement Learning}
\acrodef{pl}[PL]{Predictive Learning}
\acrodef{pf}[PF]{Particle Filtering}
\acrodef{pae}[PAE]{Predictive Autoencoder}
\acrodef{een}[EEN]{Error-Encoding Network}
\acrodef{gan}[GAN]{Generative Adversarial Networks}
\acrodef{paegan}[PAEGAN]{Predictive Autoencoding Generative Adversarial Network}

\usepackage{times}

\title{Learning to track environment state via predictive autoencoding}

\author{ {\bf Marian Andrecki} \\
Heriot-Watt and Edinburgh Universities\\
Edinburgh, United Kingdom \\
M.Andrecki@hw.ac.uk \\
\And
{\bf Nicholas K. Taylor}  \\
Heriot-Watt University\\
Edinburgh, United Kingdom \\
N.K.Taylor@hw.ac.uk\\
}

\begin{document}

\maketitle

\begin{abstract}
This work introduces a neural architecture for learning forward models of stochastic environments. The task is achieved solely through learning from temporal unstructured observations in the form of images. Once trained, the model allows for tracking of the environment state in the presence of noise or with new percepts arriving intermittently. Additionally, the state estimate can be propagated in observation-blind mode, thus allowing for long-term predictions. The network can output both expectation over future observations and samples from belief distribution. The resulting functionalities are similar to those of a Particle Filter (PF). The architecture is evaluated in an environment where we simulate objects moving. As the forward and sensor models are available, we implement a PF to gauge the quality of the models learnt from the data.
\end{abstract}

\section{INTRODUCTION}


Recent years have seen artificial agents become significantly more competent at learning how to act from unstructured data. The beginning of rapid improvements can be marked with seminal work by \cite{Mnih2015} on Atari game playing using a neural \ac{rl} agent capable of human-level performance across dozens of games without modification to its architecture. Perhaps, the most impressive case so far is \textit{AlphaZero} presented in \cite{Silver2017}, where a \textit{tabula rasa} reinforcement learner, after only couple of hours of self-interaction, surpassed top human players at Chess, Shougi and Go (all with traditions longer than 1000 years). Those and many other successes are results of the newly created overlap between the fields of deep and reinforcement learning.

Unfortunately, the super-human performance in easy-to-simulate environments, such as Chess or Atari worlds, is slow to propagate to problems embedded in the real world.
One of the key obstacles for agents learning in the real world is the fact that data gathering happens on timescales familiar to humans. The agent may need to wait for minutes to learn about consequences of its actions. Contrast this with the simulated worlds, where data gathering requires milliseconds or can be parallelised by creating many instances of the agent and environment (see \cite{Mnih2016} on asynchronous \ac{rl}).

Under such circumstances, it is important for physical agents to:
\textbf{(a)} extract more knowledge from every data point
\textbf{(b)} deliberate more about how to interact with the environment (as it is more time-expensive to fail in the real world than in the imagination) 

For both of these, a physical agent pays in the currency of computation (CPU time) hoping to decrease the number of actions until training completion (to shorten wall-clock duration of the process).

These behaviours align with the concept of \ac{pl} \cite{LeCun2016}. In this approach, the agent not only strives to select actions that maximise the expected reward, but also learns about the environment in a unsupervised manner, for example trains to predict future observations. \ac{pl} was inspired by \textit{DYNA} -- an architecture from \cite{Sutton1991}, where an \ac{rl} agent trains also a forward model of the environment and uses it for planning.

This research aims to improve the extraction of information from agents' experiences to enable planning. This can be furthered by developing new methods for learning predictive models from environmental observations.

While engineers often have the sense of a given environment themselves, it may be challenging to transfer the knowledge to the agent. Occasionally, it is difficult to express one's understanding algorithmically (e.g. how to recognise a face?) At other times abstract mathematical models are brittle and do not mesh well with the noisy and complex environments because of \textit{the frame problem}. For some of these troublesome domains it might be possible to find actionable predictive models via training of deep networks on gathered data.

When available, predictive models can be used for estimation of the environment state. Tracking (also referred to as filtering) involves two stages: \emph{prediction}, which propagates the beliefs forward in time, and \emph{update} using sensory data, which merges existing belief with the new information. This approach allows one to cope with intermittent or noisy observations. Furthermore, it is possible to quantify uncertainty of the predictions and use this information for active exploration of the environment \cite{Hero2011}. Finally, by keeping a global estimate of the environment that is only partially observable, one can combine information arriving from different sensors and locations (information fusion). This field is described in more depth in \emph{Section 2}.

The problem of automated model-building from unstructured data has been receiving increasing attention in recent years. In the domain of Atari games, impressive performance was showcased in works from \cite{Oh2015} and \cite{Chiappa2017}, where trained networks reliably predicted game states for up to hundreds of time steps. Earlier, thus smaller in scale, experiments can be found in \cite{Srivastava2015} and \cite{Lotter2015}, where learning concerned videos of bouncing and rotating objects. Predicting in highly complex environments (e.g. road traffic) was explored in both \cite{Lotter2016} and \cite{Mathieu2015}. A recent work from \cite{Henaff2017} enables sampling future observations in stochastic worlds. The recounted architectures are presented in more detail in \emph{Section 3}.

In this work, we present a neural architecture for learning predictive models of environments from high-dimensional unstructured time series data, such as videos. We show, that a trained network can be used to track the state of a non-deterministic environment in a manner comparable to \textit{Sequential Monte Carlo} methods (also known as \textit{Particle Filtering}). That is, it continually consumes observations and provides high-quality predictions of the next observation. Furthermore, the network can be run in \textit{blind} mode to generate predictions for more distant futures. The architecture can cope with observations that are intermittent or noisy. Details concerning principles of operation and implementation are expanded upon in \emph{Section 4}.

The environment used in the experiments was a simulation of moving balls. The observations were $(28 \times 28 \times 1)$ images. As the exact stochastic model was available, a Sequential Monte Carlo simulation was implemented to assess the performance of the network against a near-optimal belief tracking method. The performance of the architecture is showcased in \emph{Section 5}.

The core novelty of the method is the previously unseen combination of: \textbf{(a)} learning the forward model from data \textbf{(b)} flexibility in input/output of observations/predictions (any sequence length, intermittent observations) \textbf{(c)} ability to predict stochastic environments \textbf{(d)} ability to output both the expectation over observation and samples from the belief distribution. \emph{Section 6} discusses potential uses and plans for further developments.

\section{STATE ESTIMATION}
\subsection{AGENT AND THE WORLD}

\begin{figure}[h]
\centering
\includegraphics[width=0.45\textwidth]{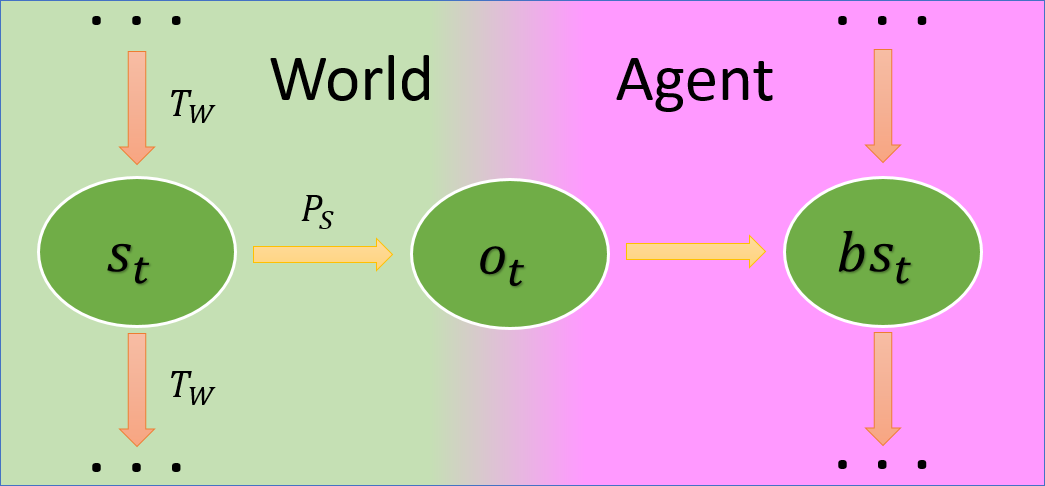}
\caption{A passive observer and its environment represented as a computational graph. The agent can learn about the environment only through the sensory channel.  \label{fig:observer}}
\end{figure}

Agents can be spawned in various environments with different tasks to complete. Figure \ref{fig:observer} is computational graph representing abstraction of a time-aware passive observer perceiving its world through an observation channel. Time flows along the downward direction. Ovals represent variables, arrows signify computational relations between those (i.e. how to compute one variable-value from another). Under the description given below, the environment is modelled as a \textit{Markov chain}.

\subsubsection{Worlds}

Left hand side relates to computations in the environment. State of a world evolves with time according to specific rules. For example, our universe is governed by particular interactions between particles. Atari and other game worlds evolve according to specific algorithm implementing them. The world state $s_t$ changes with time according to transition function $T_W$: $s_{t} = T_W(s_{t-1})$ 

The agent, implemented by computations on the right hand side\footnote{The split between agent and world is a convenient abstraction only. In truth, the agent might be embedded in its world, i.e. its state is contained in world state (e.g. humans are embedded in their environment, Atari playing algorithms are not)}, perceives its world through a sensory channel. Observation $o_t$ is a projection of environment state $s_t$ onto a, usually, much lower dimensional space defined by agents sensors: $o_t = P_S(s_t)$

The agent can never observe the underlying state directly, but by combining many observations across time it is possible to learn about environment's tendency to evolve. Agent's understanding of the environment, can be tested by examining accuracy of its predictions of future observations.

For example, a trained network from \cite{Chiappa2017} was capable of reliably predicting observations tens of time steps away. Note that this does not mean, that the network reconstructed machine code that implements Atari emulation or even that it used similar representation as Atari software developers did. It only implies, that the original environment (Atari console) and learnt emulator (neural network) are difficult to distinguished when viewed through a specific sensory channel (game screens output by the two systems). For a judge with only this channel available, these  two are largely \textit{functionally} equivalent.

\subsubsection{Axes of variation}

To better understand difficulties for predicting (and acting) in different worlds let us list some fundamental distinctions. Implications of these for learning are discussed later, when particular architectures are mentioned.

\textbf{Observability} refers to $P_S$ in a given problem. Full-observability means that the observation contains all of the information about environment state (e.g. Chess). When the world is partially-observable, it is not possible to infer its state from a single observation, thus information fusion is required.

\textbf{Observation stochasticity} describes whether $P_S(\cdot)$ is deterministic or not. Often this captures whether the observations contain noise or can be missing.

\textbf{Process stochasticity} describes whether $T_W(\cdot)$ is deterministic or not. If it is, then it is possible to simulate the environment over arbitrarily long periods of time to obtain distant reliable predictions (hence the success of predictions in Atari environment). If it is not, then it is not enough to simulate environment. One needs to propagate beliefs forward to account for different random events. The quality of prediction deteriorates with time-distance.

\textbf{Process complexity}: it is possible for the underlying environment to be completely deterministic, yet composed of so many parts or complicated rules that to the agent it appears stochastic (either because of lacking model or shortage of computational resources). Systems that are deterministic but chaotic\footnote{a small change in initial conditions leads to large divergence of outcomes} often have to be modelled as stochastic anyway.

\subsubsection{Agents}

What computations implement the agent depends on the task it is fulfilling. Still, at our level of abstraction, the agent has mental state at time $t$, call this $bs_t$ (for belief state). This state evolves with time, typically to reflect changes in the environment. This is implemented by belief time propagation and belief update with new observation (note the recursion):

\begin{equation}
 bs_{t} = T_B(bs_{t-1}, o_t)
 \label{eq:belief_update}
\end{equation}

What information is represented in the belief state depends on the problem and the environment. To give examples, an algorithm for tracking air planes with use of radar data will store information about objects of interest and discard everything else. A reinforcement learning agent may represent only those aspects that enable better estimation of expected future reward, for example colours of game objects may be often ignored. For an agent that reliably receives a rich observation at every time step (e.g. Atari player), there is little to gain from propagating beliefs forward in time -- the percept usual contains all that is of interest.

Representing the problem as a computational graph enabled us to clearly observe different challenges that arise during learning. This view translates well into design of modular neural networks. Lastly, the abstraction relates only to non-acting agents as those are studied in this work. However, it could easily be expanded to represent also reinforcement learners by adding reward signal (coming from the world to agent) and granting agent control over a fraction of world state (thus affecting how environment changes with time).

\subsection{PARTICLE FILTERING}

One of the solutions to state estimation and future prediction is \ac{pf} (also known as sequential Monte Carlo). The approach is now introduced as it was used as a (strong) benchmark to assess performance of the proposed neural architecture. For an in-depth presentation, consult \cite{Doucet2001}.

\ac{pf} maintains a probabilistic estimate over the state of interest, for example, 2D pose of a robot: $\mathbf{s} = (x, y, \theta)$). The distribution is represented by a collection of $n$ samples (\textit{particles}) -- instantiations of $\mathbf{s}$. When the robot needs to make a decision, it can operate on the knowledge of probabilistic distribution over states.

An implementation of \ac{pf} requires two core ingredients: \textbf{(a)} \text{forward model of process} -- this is $T_W$ from the earlier abstraction \textbf{(b)} \textit{sensor model} -- related to $P_S$ in the abstraction.

The propagation of beliefs involves two steps. \textbf{Prediction}: all of the samples are advanced in time with use of the forward model. This increases the uncertainty in the estimate as the forward process is stochastic. \textbf{Update}: new information from the observation is incorporated into belief. This achieved by \textit{importance sampling} of the particles using the likelihood function (particles are replicated in the proportion of their likelihood to be the origin of the observed measurement). 

\ac{pf} is a flexible method relative to its competitors. It is simple mathematically, copes well with non-linear models. Particles can represent arbitrary distributions (e.g. Kalman filter is restricted to Multinomial Gaussians). In the (unpractical) limit of infinity of particles, \ac{pf} is an optimal method for propagation of beliefs (assuming accurate models are provided).

However, all of this comes at a high computational cost. The higher the number of samples used, the more times the forward model has to be called per time step. And, usually, the more complex the problem (i.e. more non-linearities, higher dimensionality), the more particles are needed.

Still, for the experiments presented in this work, \ac{pf} is expected to perform close to optimality. Thus, it marks the top bound for performance on our problem. It can be often difficult to quantify performance of predictive neural networks. When a future observation is produced, how does one measure its accuracy? Mean squared difference between output and measured images is often uninformative. Sometimes researchers visually judge the outputs and score them for realism. In this case, we have an opportunity to compare outputs of a predictive network to visual samples from near-optimal belief tracker.

It is important to note that the predictive architectures discussed, solve a more difficult challenge than \ac{pf}. While sequential Monte Carlo relies on the forward and sensor model provided by the designer, the networks need to determine those from percepts. Additionally, the networks cope with unstructured data.

\section{PREDICTIVE NEURAL NETS}

For problems where forward models are unavailable, deep predictive architectures are becoming viable solutions. Despite the apparent difficulty, the field is advancing quickly. Atari worlds are close to solved\footnote{see videos supplied by \cite{Chiappa2017} at \href{https://sites.google.com/site/resvideos1729/}{https://sites.google.com/site/resvideos1729/}}. However, in many ways, Atari is the easiest case of the problem. The system is almost fully-observable and deterministic. 
Thus, to predict, it suffices to simulate the environment rather than maintain a distribution over the possible states. This section discusses neural approaches deployed for predicting in a range of environments.

\subsection{TOOLBOX}

\subsubsection{Predictive autoencoders}

At the high level many of the successful architectures are a combination of convolutional, recurrent and deconvolutional layers, this is sometimes termed as \ac{pae}. In essence, an input image at time $t$ is encoded in low-dimensional space via convolution. Then this encoding is combined with information about previous percepts using recurrent layers. Finally, deconvolutional layers output the prediction. The usual loss used to optimise network parameters is a mean squared difference between the actual observation and its prediction. The idea of an autoencoder relies on routing the information through a narrow passage in a neural network. At the low dimensional bottleneck, the input is represented in the encoded form. The approach can be used for image denoising, sampling from data distribution and unsupervised pretraining. Read more in \cite{Baldi2012}.

The combination of new and old information is most frequently performed with Long-Short Term Memory (LSTM) units \cite{Azzouni2017}, less so with Gated Recurrent Units (GRU) \cite{Chung2014}. In a \ac{pae} the representations formed are optimised for temporal predictions. It is possible to replace recurrent layers by concatenation multiple observations (to allow for extraction of time derivative information). However, this approach was unable to learn long-term time-dependencies in \cite{Oh2015}.

Implementations of variants of this architecture can be found in \cite{Oh2015, Chiappa2017, Henaff2017} for Atari and games, in \cite{Srivastava2015, Lotter2015, Lotter2016} for objects bouncing, rotating and natural videos.

\subsubsection{Sampling futures}

\ac{pae}s are deterministic. What happens if the underlying environment is stochastic? As argued in \cite{Mathieu2015}, the network outputs an average of possible futures (expectation over observation) and following statement from \cite{Henaff2017}, it possibly ignores low likelihood modes of the distribution (mode collapse).

This leads to multiple problems. To implement planning methods such as Markov Tree search (see \cite{browne2012}), it is important to be able to sample different possible futures, not just the maximum likelihood one. Moreover, expectations over average future percepts are often blurred and not similar to real percepts. An average of all videos from YouTube is a grey still image. A randomly sampled video probably contains a cat and is much more fun. Additionally, it is likely, that generating realistic samples requires richer representation of the environment than simply blurring the parts of frame where uncertainty arises.

\cite{Henaff2017} proposed \ac{een}, which predicts the average future for deterministic input. However, provided with source of randomness, it outputs different possible observations. Under their training scheme, the deterministic network is not required to predict every last detail of an observation correctly. Rather, once a prediction was made, a noise input to the network is tweaked to improve the match. Currently, our architecture does not implement this approach, though there appears to be much to gain.


\subsubsection{Generative adversarial networks}

\cite{Goodfellow2014} introduced a general method for sampling from high-dimensional complex distributions (e.g. images of human faces) termed \ac{gan}. The concept relies on formulating the sampling problem as a minimax game between two networks: generator of samples (\textit{G}) and discriminator (\textit{D}), who classifies samples as fake or real. The method for video prediction from \cite{Mathieu2015} uses adversarial loss to generate more realistic video frames. Our approach deploys it in similar way.

\section{PAEGAN}

\subsection{ARCHITECTURE}

\subsubsection{Overview}

We present \ac{paegan}, a neural architecture for learning forward models and tracking state using unstructured observations. When deployed, the network continually consumes observations $o$. Those can be real percepts from the environment if available or \textit{null} observations $o_{\emptyset}$ otherwise. An estimate of the environment is maintained at all times -- this value is referred to as belief state $bs_t$. At any time, it can be used to produce an expectation over possible observations $\hat{o}_{t}$, a sample observation $o'_{t}$ or a sample state $s'_t$.

To predict future percepts, the network can be run in \textit{blind} mode, with $o_{\emptyset}$ provided for belief propagation. The more distant the predictions, the more uncertain is the estimate. In observation space this is indicated by increased image blur. Examples of $o_t$, $\hat{o}_{t}$ and $o'_{t}$ are shown in Figure \ref{fig:observations}.

\begin{figure*}[h]
\centering
\includegraphics[width=0.8\textwidth]{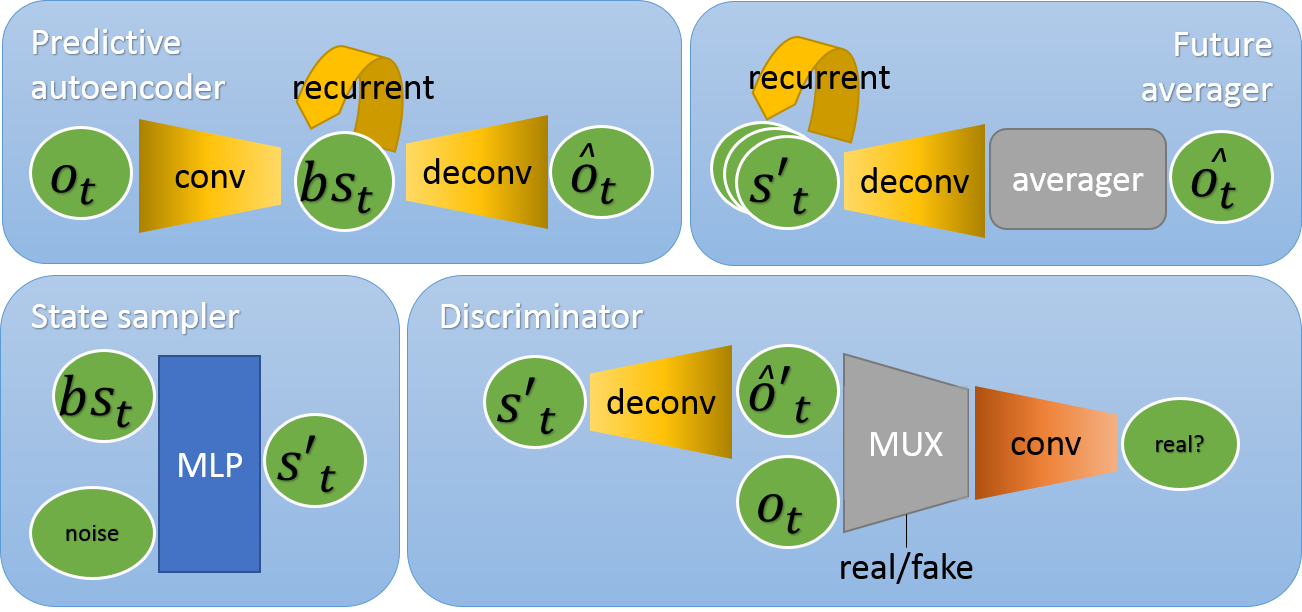}
\caption{Relations between variables expressed in terms of types of neural layers connecting them. Notice, that parts of the network are used in multiple modules, e.g. Decoder (golden \textit{deconv} trapezoid) is used in three branches. Layers are trained in stages: golden layers (PAE) first; then blue (generator) and red (discriminator) alternately; grey blocks are not trainable.}
\label{fig:architecture}
\end{figure*}

The architecture consists of modules presented diagrammatically in Figure \ref{fig:architecture}. The arrangement of the computations in the network reflects the Markovian structure of the problem. The network can also be expressed using equations from Table \ref{tab:functional}.

The implementation of the architecture in PyTorch \cite{Paszke2017} can be found in this repository\footnote{\href{https://github.com/ghostwriterm/paegan-pytorch}{https://github.com/ghostwriterm/paegan-pytorch}}. 

\begin{table}[]
\centering
\begin{tabular}{l|l}
Belief propagation & $bs_t = RNN(bs_{t-1}, Enc(o_t))$ \\ \hline
Belief propagation \\ (blind) & $bs_t = RNN(bs_{t-1}, Enc(o_{\emptyset}))$ \\ \hline
Expected observation\\ (via PAE) & $\hat{o}_t = Decoder(bs_t)$ \\ \hline
Belief sampling \\ & $s'_t = Sampler(bs_{t}, noise)$ \\  \hline
Observation sampling \\ & $o'_t = Decoder(s'_t)$ \\ \hline
Expected observation \\ (via sampling) & $\hat{o}_t = \frac{1}{n} \sum\limits_{i}^{n} Decoder(s'^{(i)}_t)$
\end{tabular}
\caption{Equations describing functional relations between variables in the architecture.}
\label{tab:functional}
\end{table}

\subsubsection{Predictive autoencoder}

\ac{pae} consists of: a convolutional \textit{encoder}, a deconvolutional \textit{decoder} and a recurrent layer (GRU). The network is trained by feeding in observations $o_{1:T}$, computing $\hat{o}_{1:T}$ (as per Table \ref{tab:functional}) and minimising MSE loss:

\begin{equation}
 \mathcal{L}_{PAE} = \sum\limits_{t=1}^{T} (o_t - \hat{o}_t)^2
\end{equation}

As the training progresses, an increasing proportion $p_{mask}$ of input observations is corrupted by overwriting with zeros (these are \textit{null} observations). Initially, $p_{mask}$ is set to $0.3$; ultimately it reaches $0.98$. With time, the network learns to fill in the missing observations. The idea is to gradually increase the difficulty of tracking. \cite{Oh2015} elaborates on importance of curriculum (gradual) training. Once the optimisation is complete, \ac{pae} allows for propagation of beliefs, updating with observations, and projection from beliefs to observations. Following \cite{Henaff2017}, \ac{pae} can be viewed as predictor of the deterministic part of the environment.

Structure of \ac{pae} allows us to reason about information content of belief state. Notice, that $bs_t$ and  $o_{\emptyset}$ (which contains no information) are sufficient to output reliable estimates of future observations. This implies, that $bs_t$ contains information about observable state, unobservable state that affects observable in future (e.g. velocities) and uncertainties about those. In other words, belief state represents a probability distribution over the environment state.

When observations are unavailable or of low quality, the belief distribution widens. At times, it is desirable to generate samples from this distribution, i.e. crisp, uncertainty-free states. The remaining modules work together to achieve this sampling capability.

Notice, that sample from belief state (i.e. $s'_t \sim bs_t$) are valid belief states themselves. Thus, those can be used as inputs for belief propagation and decoding functions.

\subsubsection{Discriminator}

\textit{Discriminator} (\textit{D}), as in \cite{Goodfellow2014}, is a convolutional network which consumes an image and outputs a classification. The network is optimised to distinguish between images from two sets: \textbf{(a)} real observations ($o_t$) \textbf{(b)} observation samples ($Decoder(s'_t)$). During training only \textit{Discriminator} layers of the are updated.

\subsubsection{Future averager}

Expected value over a distribution can be approximated by averaging value of samples from that distribution:

\begin{equation}
 \mathbb{E}[X] = \int x \ p(x) \ dx \approx \frac{1}{n} \sum\limits^{n} x \sim X
\end{equation}

Similarly, if we can generate sample states $s'_t$ from belief state $bs_t$, then the expected observation can be approximated by averaging observations that result from those samples (as per last row of Table \ref{tab:functional}). Furthermore, samples can be propagated further in time using the recurrent layer. This enables computation of future expected observations. Notice, that we already had a way to compute both of these using \ac{pae} alone.

\subsubsection{Belief state sampler}

\textit{Sampler} consumes $bs_t$ and a noise vector. Since the output is not deterministic a source of randomness is required. This module is a simple Multilayer perceptron (MLP). The output of the \textit{Sampler} is constrained by two losses:

\textbf{Generator loss} -- every sample $s'$ produced is projected onto observation space and assessed by the \textit{Discriminator}. This incentivises the \textit{Sampler} to generate samples with no uncertainty in the observable state.

\begin{equation}
 \mathcal{L}_{G} = BCE Loss(Discrimnator(Decoder(s')), \ 1)
\end{equation}

\textbf{Averager loss} -- future expected observation $\hat{o}_{t+T}$ should be the same whether computed by \ac{pae} or \textit{Observation averager}. For set of $n$ generated samples $s'^{1:n}_t$, the loss is:

\begin{equation}
 \mathcal{L}_{Av} = [Decoder(bs_{t+T}) - \frac{1}{n} \sum\limits^{n} Decoder(s'^{(i)}_{t+T})]^2
 \label{eq:constraint}
\end{equation}

Where $s'_{t+T}$ is computed from $s'_t$ using blind belief propagation (Table \ref{tab:functional}).

Due to this loos the generated samples are required to \textit{add up} to the belief state from which they were sampled. This prevents mode collapse. The overall loss for the \textit{Sampler} training is:

\begin{equation}
 \mathcal{L}_{Sampler} = \lambda_{G} \mathcal{L}_{G} + \lambda_{Av} \mathcal{L}_{Av} 
 \label{eq:losses}
\end{equation}

\textit{Sampler} and \textit{Discriminator} are trained interchangeably while \ac{pae} is held constant. 

\subsubsection{Design principles}

During the design we focused more on the overarching architecture than on the low level hyperparameters (such as particular activation functions or inclusion of normalising layers). \ac{paegan}'s structure aims to reflect the underlying \textit{Markov chain} estimation problem. We rely on controlling the flow of information through the network to enforce particular information content in the learnt representation. For example, during \ac{pae} training, observations are intermittent to incetivise propagation of information through time. Additionally, we attempt to force specific representations (e.g. uncertainty-free samples) by constraining intermediate outputs of the network with particular loss functions.

Good performance in experiments despite little focus on low level parameters, possibly indicates generality of the architecture. We used generic and simple neural modules: encoder and decoder are generic, 3-layered convolutional networks with ReLU activations, RNN is 1-layer of Gated Recurrent Units, \textit{Discriminator} is taken from DCGAN \cite{Radford2015}. Batch normalisation is not used apart from DCGAN. Belief state is represented by 256 numbers at the recurrent layer. For all optimisation we used \textit{Adam} with learning rate $0.0002$.

The values of $\lambda_{G}$ and $\lambda_{Av}$ that ensure good performance depend on the data used for training (e.g. we chose $\lambda_{G}=1, \ \lambda_{Av}=500 $). Overall, the two loss functions should have similar contribution to the training.

\section{EXPERIMENTS}
\subsection{SETUP}

\ac{paegan} was tested in a setup similar to \cite{Lotter2015}. The environment is a simulation of moving balls. Depending on the settings, the balls bounce from each other (and walls) or phase through. Targets are moving with near constant velocity, with noise added. The velocities and positions of the balls are continuous variables. The simulator produces observations from the environment state. These are $28 \times 28 \times 1$ images, which represent information about positions of targets, see Figure \ref{fig:observations}. Those can be corrupted with noise (random shifts to target positions) or completely accurate. Using terms from Section 2, the environment is stochastic and partially-observable via intermittent and noisy unstructured percepts.

The network was trained using purely time series of observations. For memory-management purposes, the duration of training episodes was $100$ time steps. At test time, a fixed number of initial observations ($8$) was provided with \textit{null} observations following. The state tracking was assessed by inspecting MSE of predicted observations over $160$ time steps. This allowed to quantify retention of the information gained from percepts. Tracking was assessed also for observations arriving randomly (e.g. with $5\%$ chance per time step). This emulates state tracking when the sensor is not reliable or decisions need to be made without new percepts. We also extract sample observations and assess their realism and variability. Multiple training and evaluation runs were completed for different dynamics of the simulator. Scenarios included one or many balls moving, in some cases interacting via collisions.

As both the forward and sensor models are known exactly, it was possible to implement a \ac{pf}. Generally, when accurate stochastic models are provided, the \ac{pf} is an optimal state estimator in the limit of infinity samples used. In our case, since the process at hand was not strongly non-linear or high-dimensional, $1000$ particles assured good performance. The data fed was a structured measurement of position for every target in the scene.

Because \ac{pf} maintains a belief distribution over state, we can generate the expected observation conditional on that belief (using the sensor model). This means \ac{paegan} and \ac{pf} can be compared on the same metric of image prediction. Note, that \ac{pf} is a very strong benchmark and solves an easier problem: state estimation with models provided. \ac{paegan} learns models from the data and then performs predictions.

The implementation of experiments is also available in the project repository.

\subsection{RESULTS}

\begin{figure*}[h]
\centering
\includegraphics[width=0.8\textwidth]{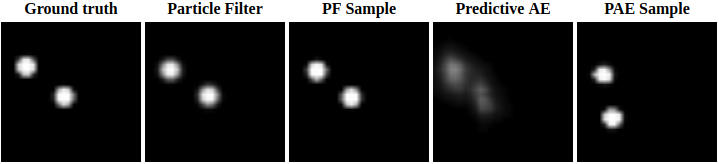}
\caption{Observation estimates produced by \ac{pf} and \ac{paegan} after not observing the environment for $40$ time steps. From the left: an observation with two balls (prediction target); expectation and sample observation from \ac{pf}; finally, corresponding outputs from \ac{pae}. Image blur signifies uncertainty; it is not present in samples.}
\label{fig:observations}
\end{figure*}

\begin{figure*}[h]
\centering
\includegraphics[width=0.8\textwidth]{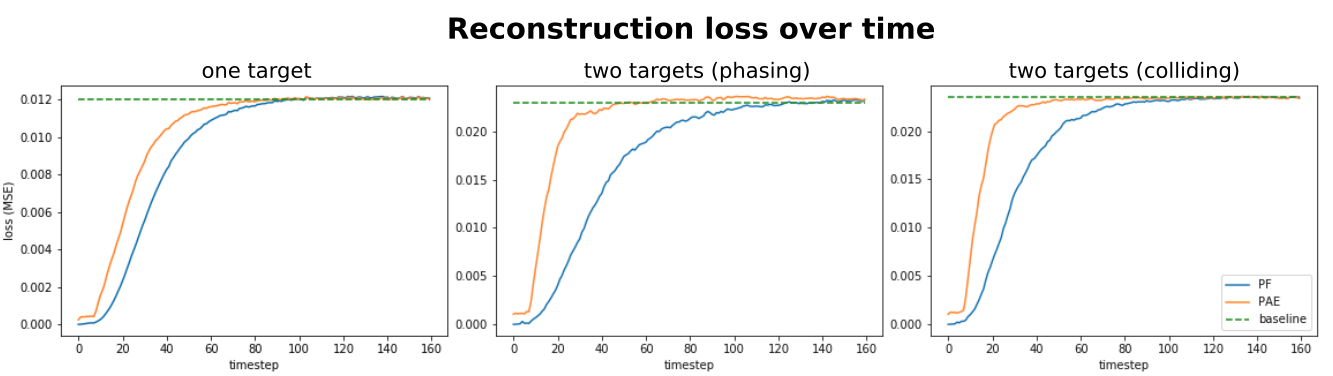}
\caption{Plots show reconstruction loss for predictions in increasingly complex environments (left to right). \ac{pf} provides more reliable reconstructions than \ac{paegan}. The difference is especially clear for more environment physics (right-most plot) . Green dotted line is an uninformed baseline -- best level achievable without access to any recent observations (by outputting an average observation). }
\label{fig:plots}
\end{figure*}

The predictions output by \ac{paegan} are similar to those of \ac{pf} (albeit more uncertain). Consider Figure \ref{fig:observations}, but it is best to see videos here\footnote{\href{https://docs.google.com/document/d/1Tz8UO0sDELuTHgz6CkDlc_9VIc6eCv1i3xRA2Ba0leI/edit?usp=sharing}{GoogleDoc}}. This favours the hypothesis from \cite{Mathieu2015}, that a deterministic predictive network (\ac{pae}) approximates the expectation over the target value. Figure \ref{fig:plots} shows how reconstruction error gradually increases when observations are not available. As should be expected, \ac{pf} is more successful at tracking the environment state. This is because \ac{pae} did not infer the forward model perfectly. The difference is especially apparent in more complex environments as models of those are more difficult to extract from data (e.g. inherently chaotic simulation of collisions). Notice, that neither method does worse than the uninformed baseline (dotted green line). At this error level, the estimate reaches maximum uncertainty.

Nonetheless, \ac{paegan} learns to represent and recursively propagate information about the environment, effectively learning the underlying physics. Importantly, it offers not only predictions, but also associated uncertainties. The representation contains information also about unobservable state of the environment -- velocities of targets need to be inferred for successful predictions.

While this is not shown here (see the videos page linked earlier), \ac{paegan} also reliably tracks state when observations arrive infrequently or contain noise. This process is referred to as \textit{smoothing} and often is added to systems (e.g. in form of \ac{pf}) to enable decision-making when recent observations are likely to be unavailable.

Observation samples produced are realistic and appear to be taken from the belief distribution (see Figure \ref{fig:observations} and videos). Normally, it is difficult to assess whether samples cover an entire distribution. Realism and diversity are typically assessed by the operators. In this case, ensuring that many samples together \textit{add up} to the distribution of origin is an explicit training goal (Equation \ref{eq:constraint}). The quality of the samples decreases when belief state is at maximum uncertainty: it appears that not the entire state space is sampled (mode collapse).

The duration of training for the \ac{pae} module strongly depends on the complexity of the environment. Learning collision kinematics of two balls required about $10\times$ more parameter updates than training for a single moving target. With prolonged optimisation, the risk of overfitting was higher. If provided with a small data set, the network tended to memorise the trajectories instead of learning the underlying models. This can be easily detected by validating the model on held out data.

Performance of the architecture is bounded by the quality of \ac{pae} predictions. For many problems, predicting the average future outcome is much easier than generating samples of the future. This means that, in general, the representations formed by a \ac{pae} are not necessarily well matched for sampling. It is possible to improve this by combining our approach with work from \cite{Henaff2017}, where noise input to the network is used as an explanatory variable of the observations. In our particular experiments, the average outcome is indeed informative about various consequences.

Concurrent training of \textit{Sampler} and \textit{Discriminator} displays lack of stability representative of \ac{gan} approaches. In our case, optimising the \textit{D} network less frequently than \textit{G} improved the quality of generated samples. It could be argued that, since \textit{G} is more constrained (see Equation \ref{eq:losses}), it demands more search through the parameters.

\section{DISCUSSION AND FURTHER WORK}

\ac{paegan} is a combination of existing attempts at prediction of unstructured time series using neural networks. We present it in the frame of \textit{Markov chain} estimation. From a black-box perspective, \ac{paegan} can be seen as a \textit{trainable} Particle Filter for unstructured data. The recursive structure enables flexibility of inputs and outputs. The observations can be noisy and intermittent, the outputs include expectation over future observations, samples of future observations and belief states (as a neural representation).

The state tracking method can be directly connected to neural reinforcement learning with multiple expected advantages. Firstly, the representations useful for predictions are likely relevant for related tasks, such as policy search and value estimation. Therefore, when training \ac{paegan} in an unsupervised manner, the agent is able to extract more information from its experiences. Secondly, a forward model of the environment enables planning. Thirdly, when observations are unreliable (e.g. server latency in \textit{OpenAi Universe}), then the agent can rely on the estimate of environment state. Lastly, agents in largely unobservable worlds (like ours) need to combine information across time, sensors and locations. Similar problems (albeit with structured data) are today solved using information fusion methods, an example of which is the Particle Filter. It is foreseeable that a descendant of \ac{paegan} could perform a similar function. These advantages are especially valuable for agents embedded in the real world (e.g. robotic systems), which need to learn from limited data. The direct next steps for the architecture are to apply it to more complex environments in a reinforcement learning framework (e.g. Atari games) with focus on data-efficient learning and planning.

Another research avenue is the improvement of quality of the predictions. Currently, the network encodes the environment estimate using the state of a recurrent layer. This means that a single propagation operation needs to track possibly multiple modes. Alternatively, it is possible to express knowledge about state as a collection of samples from a given belief distribution (as is done in Particle Filter). In some situations, it might be valuable to represent the state estimate as a combination of belief states by multiple instantiations of \ac{paegan}s. Humans intuitively predict the future using a mixture of sampling and uncertainty representation. Sufficiently distinct events are evaluated as different branches of the future each with their own associated uncertainties. For example: 'I might go to work either by car or bus' -- two distinct scenarios. However, once we zoom in on a particular path, we uncover uncertainties particular to this case (e.g. how much do I expect to pay for a bus ticket). Implementing a similar prediction mode using \ac{paegan} might bring two gains. Firstly, improved quality of predictions, as a single instantiation of the network needs to propagate only unimodal beliefs. Secondly, improved computational efficiency for approaches such as Monte Carlo tree search, as the branching could be performed only at the significant events when the belief distribution diverges into multiple modes. In such a view, the nodes of the tree would be belief states rather than state samples.


\newpage
\bibliographystyle{apalike}
\bibliography{paper}

\end{document}